\documentclass[sigconf]{acmart}
\usepackage{multirow}
\usepackage{enumitem}
\AtBeginDocument{%
  }

\begin{document}

\title{MDE-Edit: Masked Dual-Editing for Multi-Object Image Editing via Diffusion Models}

\author{Hongyang Zhu}
\affiliation{%
\institution{School of Computer Science and Information Engineering, Hefei University of Technology}
\streetaddress{}
\city{Hefei}
\country{China}
\state{}}
\email{hongyangzhu@mail.hfut.edu.cn}

\author{Haipeng Liu}
\authornote{Haipeng Liu is the corresponding author}
\affiliation{%
\institution{School of Computer Science and Information Engineering, Hefei University of Technology}
\streetaddress{}
\city{Hefei}
\country{China}
\state{}}
\email{hpliu_hfut@hotmail.com}

\author{Bo Fu}
\affiliation{%
\institution{School of computer and artificial intelligence, \\Liaoning Normal University}
\streetaddress{}
\city{Dalian}
\country{China}
\state{}}
\email{fubo@lnnu.edu.cn}

\author{Yang Wang}
\affiliation{%
\institution{School of Computer Science and Information Engineering, Hefei University of Technology}
\streetaddress{}
\city{Hefei}
\country{China}
\state{}}
\email{yangwang@hfut.edu.cn }
\begin{abstract}
Multi-object editing aims to modify multiple objects or regions in complex scenes while preserving structural coherence. This task faces significant challenges in scenarios involving overlapping or interacting objects: (1) Inaccurate localization of target objects due to attention misalignment, leading to incomplete or misplaced edits; (2) Attribute-object mismatch, where color or texture changes fail to align with intended regions due to cross-attention leakage, creating semantic conflicts (\textit{e.g.}, color bleeding into non-target areas). Existing methods struggle with these challenges: approaches relying on global cross-attention mechanisms suffer from attention dilution and spatial interference between objects, while mask-based methods fail to bind attributes to geometrically accurate regions due to feature entanglement in multi-object scenarios. To address these limitations, we propose a training-free, inference-stage optimization approach that enables precise localized image manipulation in complex multi-object scenes, named MDE-Edit. MDE-Edit optimizes the noise latent feature in diffusion models via two key losses: Object Alignment Loss (OAL) aligns multi-layer cross-attention with segmentation masks for precise object positioning, and Color Consistency Loss (CCL) amplifies target attribute attention within masks while suppressing leakage to adjacent regions. This dual-loss design ensures localized and coherent multi-object edits. Extensive experiments demonstrate that MDE-Edit outperforms state-of-the-art methods in editing accuracy and visual quality, offering a robust solution for complex multi-object image manipulation tasks.
\end{abstract}

\renewcommand\footnotetextcopyrightpermission[1]{}
\settopmatter{printacmref=false} 

%

\begin{teaserfigure}
  \includegraphics[width=\textwidth]{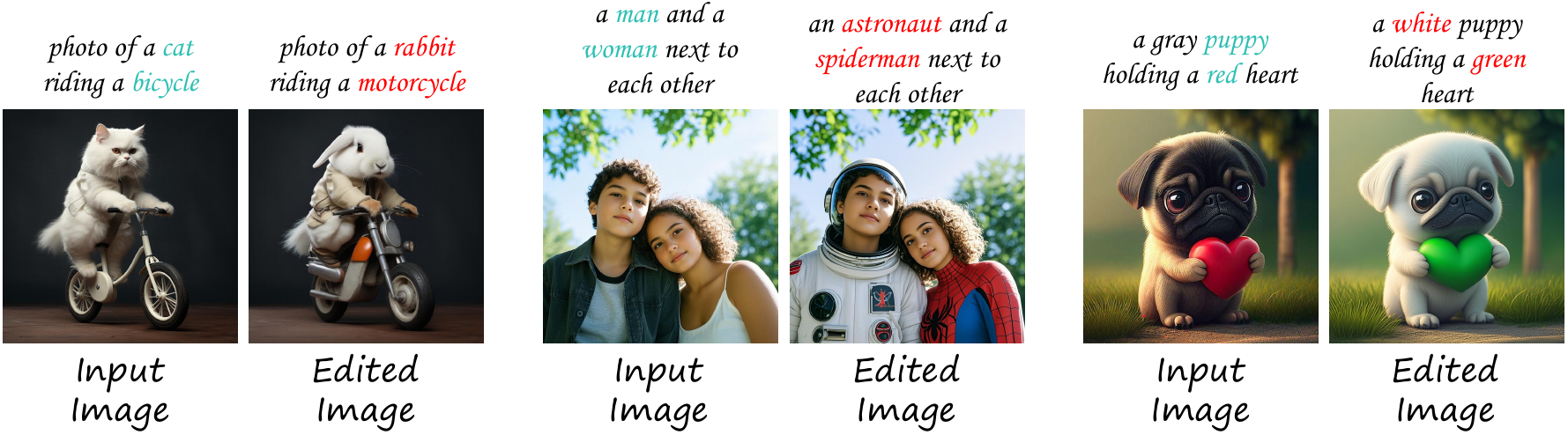}
  \caption{Representative results of MDE-Edit on diverse image editing tasks. For each example, the left image displays the original input, and the right image shows the edited output. The text above the image represents the source prompt and target prompt respectively. Our approach achieves precise multi-object modifications, addressing localized challenges such as recoloring, object replacement, and maintaining structural consistency.}
  \label{fig:1}
\end{teaserfigure}


\maketitle

\begin{figure*}
    \centering
    \includegraphics[width=0.8\linewidth]{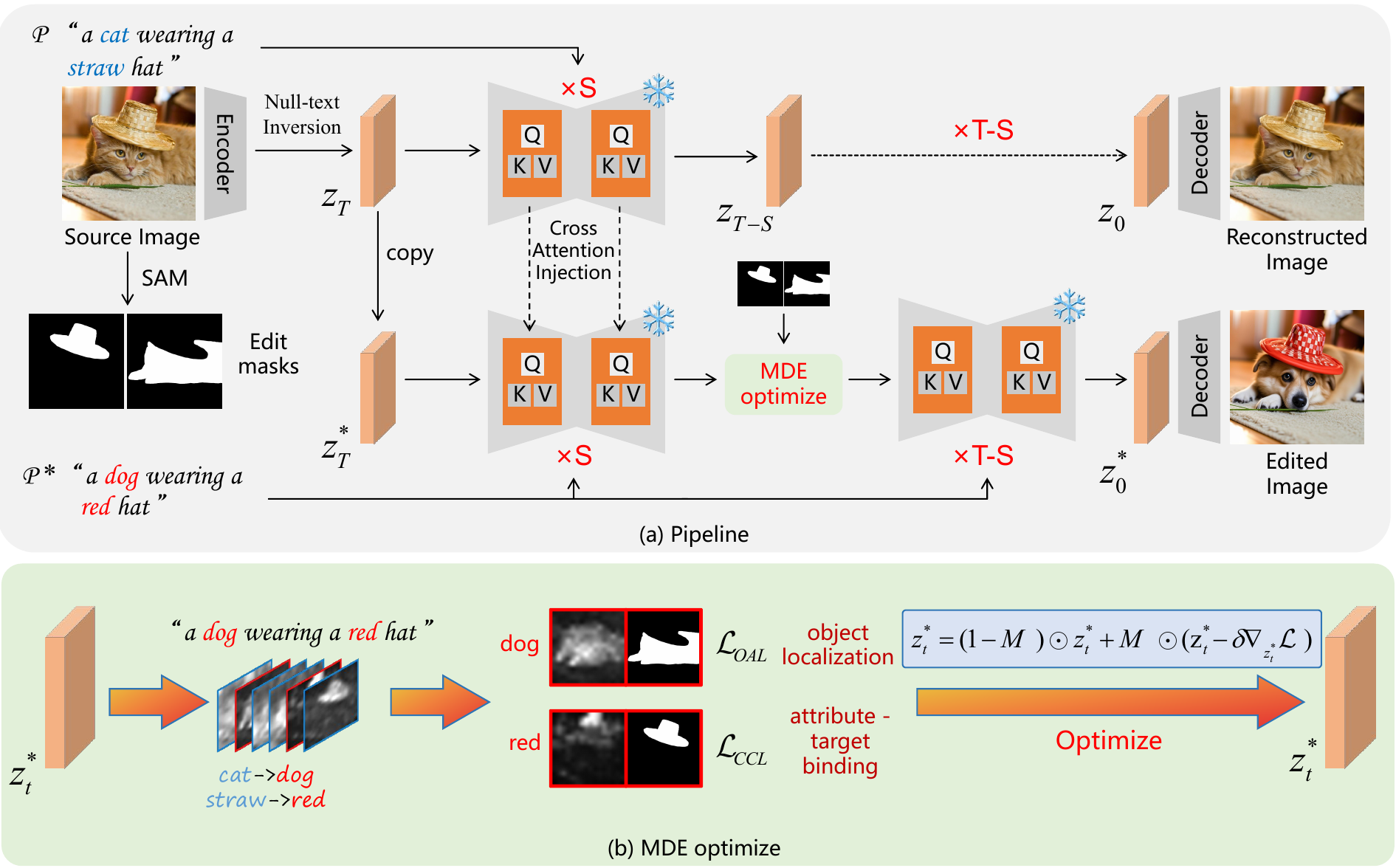}
    \caption{Illustration of our proposed MDE-Edit framework. (a) The proposed MDE-Edit framework consists of two main branches: the upper row depicts the reconstruction branch, while the lower row represents the editing branch. Within the editing branch, the noise latent feature undergoes optimization through the MDE process. This optimization mechanism facilitates precise alignment with the specified edit prompts, namely “dog” and “red hat”, while simultaneously maintaining focus within the designated edit region masks.
    (b) Illustration of MDE optimization. OAL and CCL respectively handle the positioning of the object and the binding of attributes to the object.}
    \label{fig:2}
\end{figure*}

\section{Introduction}
With significant progress in large-scale text-to-image (T2I) diffusion models—including Stable Diffusion\cite{rombach2022high}, DALL·E\cite{ramesh2022hierarchical}, and Imagen\cite{saharia2022photorealistic}—researchers have observed their exceptional ability to generate high-quality, diverse images. These models have also catalyzed advancements in multiple downstream tasks, such as text-guided image editing\cite{brooks2023instructpix2pix, hertz2022prompt, meng2021sdedit, sheynin2024emu, hertz2023delta, nam2024contrastive, titov2024guide, goel2023pair,kawar2023imagic,shi2024dragdiffusion,li2023layerdiffusion} and image inpainting\cite{liu2024structure, liu2022delving}, attracting substantial attention from both academia and industry.

The application of diffusion models in text-guided image editing has experienced a surge in interest, evidenced by the significant number of research publications in this field over recent years. Existing methods primarily focus on single-object editing in simple scenarios. However, real-world images typically consist of multiple objects with complex spatial relationships, and users often require the ability to edit these objects simultaneously. As shown in Fig. \ref{fig:1}, the images contain at least two interacting objects, and we perform corresponding edits on them while preserving contextual coherence.

In complex multi-object scenes, objects may overlap, interact, or exhibit intricate spatial arrangements. Performing fine-grained editing on multiple objects while preserving background elements presents substantial challenges. The key difficulties manifest in two aspects: (1) Inaccurate target localization due to attention misalignment—mutual interference between objects prevents complete attention matching, leading to incomplete or misplaced edits; (2) Attribute-object mismatch caused by cross-attention leakage, where entangled features of overlapping objects in latent space result in unintended propagation of color/texture modifications to non-target regions, creating semantic conflicts such as color bleeding or texture misalignment.

Previous research in text-guided image editing has made efforts to overcome these challenges. Early methods like Prompt-to-Prompt (P2P)\cite{hertz2022prompt} and Plug-and-Play (PnP)\cite{tumanyan2023plug} utilize cross-attention mechanisms to maintain structural coherence during the editing process. P2P attempts to map the cross-attention of the source image to the target image based on the text prompts, while PnP uses pre-trained models sequentially to update the image. However, in multi-object scenarios, they often suffer from attention misalignment or leakage. When multiple objects are present, the cross-attention mechanism may not accurately focus on the specific object to be edited, leading to incorrect modifications in other parts of the image.
Object-aware Inversion and Reassembly (OIR)\cite{yang2023object} is a mask-based method specifically designed for multi-object editing in text-driven image diffusion models. The approach employs a "disassembly then reassembly" strategy: it first identifies and processes each target object separately using optimized inversion steps determined by a novel search metric, avoiding concept mismatch during editing. After modifying the specified regions, OIR integrates the edited objects with non-edited areas to produce a coherent final image. However, a key limitation of OIR is its reliance on clear segmentation masks for precise editing, which becomes unreliable in scenarios where objects overlap or exhibit complex spatial interactions.

In this paper, we introduce a novel framework for multi-object image editing using diffusion models to address long-standing challenges in precise object manipulation. Our method builds upon latent diffusion models (LDMs)\cite{rombach2022high} and introduces a Cross-Attention Control mechanism that extracts and modulates cross-attention maps from pre-trained text-to-image diffusion models. This enables fine-grained control over specific regions or objects while preserving the original image’s structural integrity. By manipulating these attention maps, we ensure generated content aligns with target text prompts while minimizing interference between objects in complex scenes. In particular, we propose a \textbf{M}asked \textbf{D}ual \textbf{E}dit strategy that combines Object Alignment Loss (OAL) and Color Consistency Loss (CCL). OAL leverages implicit segmentation from cross-attention layers to maintain accurate object positioning and scale, addressing issues like attention dilution and spatial variance. CCL enforces color consistency through selective attention boosting, gradient amplification, and spatial focus mechanisms, ensuring color changes are confined to masked regions. This dual-loss architecture separates structural and appearance editing, enabling semantically accurate and visually coherent multi-object modifications.

Our contributions can be summarized as follows:
\begin{itemize}[topsep=1pt]
\item We focus on complex scenes where multiple objects exhibit complex spatial relationships or intricate interactions, which challenge existing methods in maintaining precise attribute-object associations and localized edits.
\item We introduce a Masked Dual-Edit strategy that combines Object Alignment Loss (OAL) and Color Consistency Loss (CCL) to separately control object structure and appearance, ensuring that edits are both semantically accurate and visually consistent.
\item We demonstrate the effectiveness of our approach through extensive experiments on custom benchmark datasets, showing significant improvements in both qualitative and quantitative metrics compared to state-of-the-art methods. Our code can be accessed in the supplementary material.
\end{itemize}

\vspace{-4pt} 
\section{Related Work}
\noindent \textbf{Text-to-image diffusion models.}
Diffusion models have emerged as a powerful tool for image generation, particularly in text-to-image (T2I) applications. By conditioning the denoising process on text prompts, these models generate images that align with textual descriptions. However, challenges persist in multi-object generation, such as missing objects or attribute mismatches. Recent work, like Attend-and-Excite\cite{chefer2023attend}, addresses these issues by optimizing cross-attention during inference to better focus on subject tokens. The inherent constraints of diffusion models in multi-object generation frequently result in inadequate performance for editing tasks requiring simultaneous modifications of multiple entities. Knowledge distillation techniques\cite{wang2024unpacking, qian2023adaptive, qian2023rethinking} have also been explored to improve efficiency, leveraging lightweight models to approximate the behavior of larger diffusion models without sacrificing precision. These developments highlight the ongoing efforts to enhance the robustness and accuracy of diffusion models in complex editing tasks.

\noindent \textbf{Text-guided Image Editing.}
Text-guided image editing\cite{brooks2023instructpix2pix, hertz2022prompt, meng2021sdedit, sheynin2024emu, hertz2023delta, nam2024contrastive, titov2024guide, goel2023pair,kawar2023imagic,shi2024dragdiffusion,li2023layerdiffusion} has evolved through several key phases, driven by advancements in generative models and semantic understanding. Early approaches combined GANs\cite{goodfellow2020generative} with CLIP\cite{radford2021learning}, as seen in methods like StyleCLIP\cite{patashnik2021styleclip}, which leveraged CLIP’s aligned image-text representations to enable semantic-driven manipulations. However, GANs faced limitations in inversion fidelity, often compromising reconstruction quality when mapping real images to latent spaces. The emergence of diffusion models addressed these challenges, offering superior inversion and denoising capabilities. Initial diffusion-based techniques, such as SDEdit\cite{meng2021sdedit}, introduced noise to the input image and denoised it under text guidance, balancing realism and edit faithfulness but struggling with localized changes. Subsequent methods like Blended Diffusion\cite{avrahami2022blended} and DiffEdit\cite{couairon2022diffedit} incorporated masks to restrict edits to specific regions, automating mask generation and refining inversion strategies to preserve unedited areas. These methods can effectively keep unedited areas outside the mask intact. However, they often cause significant structural changes within the edit regions, leading to inconsistencies in multi-object scenes where contextual coherence is critical.

Structural coherence became a focus with attention-based approaches: Prompt-to-Prompt (P2P)\cite{hertz2022prompt} manipulated cross-attention maps to align generated content with text prompts, while Plug-and-Play (PNP)\cite{tumanyan2023plug} injected source image features into the denoising process for direct natural image editing. Inversion techniques also advanced, with Null-text Inversion\cite{mokady2023null} optimizing null embeddings for precise reconstruction. Despite progress, multi-object editing posed challenges—simultaneously editing multiple regions, maintaining contextual consistency, and avoiding error propagation. Delta Denoising Score (DDS)\cite{hertz2023delta} refines text-guided image editing by subtracting noisy gradients using a reference image-text pair, while Contrastive Denoising Score (CDS)\cite{nam2024contrastive} further enhances DDS with CUT loss via self-attention features from latent diffusion models to enforce structural consistency between source and edited images. Guide-and-Rescale (GnR)\cite{titov2024guide} introduces a tuning-free real-image editing method leveraging self-guidance and noise rescaling. Object-aware Inversion and Reassembly (OIR)\cite{yang2023object} is specifically designed for multi-object editing but struggles to disentangle spatially entangled objects, limiting its effectiveness in scenes with intricate object interactions.

\section{Preliminary}
\noindent \textbf{Latent Diffusion Models.}
We apply our method to the open-source Stable Diffusion (SD)\cite{rombach2022high}, which generates high-quality images from noise. SD operates in three key stages. First, an auto-encoder is trained to map an image \(x\) into a lower-dimensional latent representation \(z = \mathcal{E}(x)\), where \(\mathcal{E}\) denotes the encoder. Second, the diffusion process is applied to the latent representations \(z\). The forward diffusion process gradually adds Gaussian noise, transforming \(z\) into a noisy latent \(z_{t}\), where the time step \(t\) is uniformly sampled from \(\{1, \cdots, T\}\). Then, a neural network learns to denoise \(z_{t}\) into new samples \(z_{0}\) iteratively. The denoising stage is called the reverse process. The conditional reverse process is defined as:
\begin{equation}
\mathbb{E}_{\mathcal{E}(x),y,\epsilon\sim N(0,1),t} \left[ ||\epsilon-\epsilon_{\theta}(z_t,t,\tau_{\theta}(y))||_2^2 \right],
\label{eq:1}
\end{equation}
where \(\epsilon\) is Gaussian noise sampled from a standard normal distribution; \(\epsilon_{\theta}\) is a parameterized denoising network (\textit{i.e.}, a U-Net in SD); \(\tau_{\theta}\) is a frozen CLIP text encoder and \(y\) represents text prompts. \(\epsilon_{\theta}\) is conditioned on \(z_{t}\), the time step \(t\), and the encoded conditioning information \(\tau_{\theta}(y)\). During sampling, \(z_{T}\) is randomly sampled from a standard Gaussian distribution and iteratively denoised by \(\epsilon_{\theta}\) to produce \(z_{0}\). Finally, a decoder \(\mathcal{D}\) reconstructs \(z_{0}\) into an image \(\hat{x}\) of a specific size, mathematically expressed as \(\hat{x} = \mathcal{D}(z_0)\).

\noindent \textbf{Cross-Attention Mechanism.}
SD introduces the cross-attention mechanism to guide image synthesis using text prompts. The cross-attention map (CA map) \( A \) is defined as:
\begin{equation}
A_l = \text{softmax} \left( \frac{QK^T}{\sqrt{d_k}} \right),
\label{eq:2}
\end{equation}
where \( A_l \) represents the cross-attention map at layer \( l \), \( d_k \) is the dimensionality of the keys. In SD, \( Q \) (query) is mapped from the intermediate image representation, while \( K \) (key) is linearly projected from the conditioning text embedding. A pretrained CLIP text encoder is used to encode the text prompt into a sequential text embedding.

\noindent \textbf{Null-Text Inversion (NTI).}
Real image editing requires reversing the corresponding  \(z_0\)  back to \(z_T\). A straightforward DDIM inversion\cite{song2020denoising} method, in theory reversible with infinitesimally small steps, tends to accumulate reconstruction errors in the denoising process, particularly due to classifier guidance. In Null-Text Inversion\cite{mokady2023null}, DDIM Inversion is first applied to the original image to obtain the inversion trajectory. Then, during the sampling process, the null-text embedding is fine-tuned to reduce the distance between the sampling trajectory and the inversion trajectory so that the sampling process can
reconstruct the original image. The advantage of this approach
is that neither the U-Net weights nor the text embedding are
changed, so it is possible to improve reconstruction performance
without changing the target prompt set by the user.

\begin{figure}
    \centering
    \includegraphics[width=0.9\linewidth]{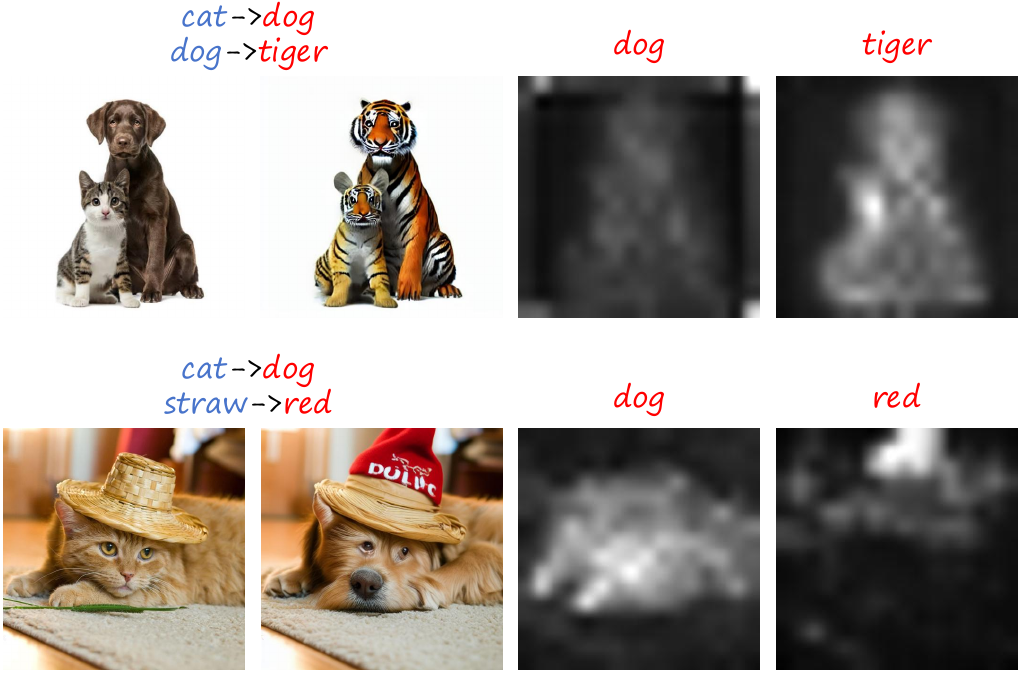}
    \caption{Multi-object editing failure cases. Revealing challenges in maintaining coherent attribute-object associations and preventing unintended hybrid outputs.}
    \label{fig:3}
\end{figure}

\section{Methodology}
Our work can be summarized as follows. (1) Attention injection to maintain spatial coherence in object layout (Sec \ref{sec:4.1}). (2) Masked Dual-Edit via Attention with Object Alignment Loss (OAL) and Color Consistency Loss (CCL) to address localization and attribute issues (Sec \ref{sec:4.2}). (3) Latent variable optimization using the combined OAL and CCL losses to control object structure and attributes while preserving image integrity (Sec \ref{sec:4.3}).

\noindent \textbf{Problem Statement:}
Multi-object image editing with diffusion models faces critical challenges in maintaining precise attribute-object associations, especially when handling concurrent object transformations. As illustrated in Fig. \ref{fig:3}, two distinct failure modes emerge: (1) In the "cat→dog, dog→tiger" example (upper), the presence of the tiger affected the dog's attention map, resulting in the absence of the dog in the edited image. (2) The "cat→dog, straw→red" case (lower) demonstrates Attribute-Object Mismatch, where the color modification intended for the straw improperly affects adjacent regions, creating inconsistent artifacts.

These failures reveal two fundamental issues:

\noindent (1) \textbf{Inaccurate Localization}:
When multiple objects are present in an image, their attention maps interfere with each other. Without proper constraints, this can lead to inaccurate or incorrect localization (Fig. \ref{fig:3}, upper).

\noindent (2) \textbf{Attribute-Object Mismatch}:
Attribute-object mismatch occurs when color or texture changes fail to align with intended regions due to cross-attention leakage. This causes the modifications to spread to non-target areas, resulting in semantic conflicts such as color bleeding into adjacent regions (Fig. \ref{fig:3}, lower).

The root cause lies in the inability of standard cross-attention maps to simultaneously achieve three critical objectives: precise spatial localization of edits, preservation of non-target regions, and maintenance of coherent attribute-object associations. Current approaches either focus on global attention control (losing spatial precision) or employ simple masking (failing to handle complex object interactions), highlighting the need for a more sophisticated attention manipulation framework.

\noindent \textbf{Overview of MDE:}
Let \( I \) be a real image, we first employ Null-text inversion\cite{mokady2023null} to encode it into the noise latent feature \( z_T \).
The conditions for this multi-object editing scenario include a source prompt \(p\) and a target edit text prompt \(p^{*}\). For each object to be edited, there is a corresponding mask \(S_{i}\) and an associated editing text prompt \(T_{i}\). To address the challenges mentioned above, our MDE-Edit framework introduces a masked dual-editing strategy. As depicted in Fig. \ref{fig:2}, the framework consists of two parallel branches. The reconstruction branch retains cross-attention (CA) for shared tokens to preserve structure, while the editing branch injects new CA for target tokens within masks. Merging these maps retains shared attention and prioritizes new tokens in masks, suppressing cross-attention leakage for structural consistency and edit specificity.
Subsequently, we introduce a masked dual-loss framework to optimize \( \mathbf{z}_t^* \), comprising two core components: the Object Alignment Loss (OAL) aligns cross-attention maps with segmentation masks to enforce spatial confinement of edits, ensuring geometric fidelity to target object boundaries and preventing unintended modifications outside masked regions; the Color Consistency Loss (CCL) enhances attention to target attribute tokens (\textit{e.g.}, color, texture) within masks while suppressing interference from shared semantic tokens, ensuring semantic coherence for attribute-specific modifications and minimizing cross-object leakage. Together, these losses decouple structural localization and appearance editing, enabling precise, context-aware multi-object edits.

\subsection{Attention Injection} \label{sec:4.1}
As illustrated in Fig. \ref{fig:2}, to preserve the structural information of the original image during editing, the cross-attention (CA) maps of common tokens from the reconstruction branch are injected into the editing branch at diffusion step \( t \), resulting in the mixed CA maps \(\hat{A}_t\). This process is defined as:
\begin{equation}
\text{Inject}(A_t, A^*_t) :=
\begin{cases}
(A_t)_j & \text{for tokens shared between } p \text{ and } p^*, \\
(A^*_t)_i & \text{for new tokens introduced in } p^* ,
\end{cases}
\label{eq:3}
\end{equation}
where \( A_t \) represents the CA maps from the reconstruction branch, and \( A^*_t \) represents the CA maps from the editing branch. This approach ensures that:
The attention values for tokens shared between the source prompt \( p \) and the target edit prompt \( p^* \) are retained from the reconstruction branch.
The attention values for new tokens introduced in the target edit prompt \( p^* \) are taken from the editing branch.

By combining these, the method maintains the structural consistency of the original image while incorporating the new content introduced during editing.

\subsection{Masked Dual-Edit via Attention} \label{sec:4.2}
In complex multi-object scenes (\textit{e.g.}, Fig. \ref{fig:3}), text-to-image diffusion models struggle to maintain coherent attribute-object associations. For instance, editing “straw hat” to “red hat” on a cat (Fig. \ref{fig:3} lower) often results in color leakage to non-masked areas due to cross-attention spreading. Similarly, transforming a “cat” to a “dog” while changing a “dog” to a “tiger” (Fig. \ref{fig:3} upper) frequently produces hybrid objects because of attention misalignment. These failures highlight the need to separately control two editing dimensions: \textbf{object replacement} (\textit{e.g.}, cat→dog) and \textbf{appearance modification} (\textit{e.g.}, straw→red). To address this, we propose a Masked Dual-Edit approach with two specialized losses:
\begin{itemize}
    \item \textbf{Object Alignment Loss (OAL)}: Ensures geometric accuracy of object replacement using per-object masks \( S_i \).
    \item \textbf{Color Consistency Loss (CCL)}: Enforces color coherence within regional masks \( S \) while suppressing interference from shared tokens.
\end{itemize}

The dual-loss framework synergizes structural preservation (OAL) and appearance control (CCL), addressing the two core challenges outlined in the Abstract: inaccurate localization and attribute leakage. This mirrors the "disassembly then reassembly" philosophy of OIR\cite{yang2023object} but extends it to complex multi-object scenarios by replacing explicit mask-based disassembly with implicit attention-guided feature disentanglement.

\subsubsection{Object Alignment Loss (OAL)}

The Object Alignment Loss (OAL) leverages implicit segmentation maps extracted from cross-attention layers to align generated content with input masks. We find that the attention
maps provide meaningful localisation information, but only when they are averaged across different attention heads and feature layers.
\begin{equation}
\hat{S}_i = \frac{1}{L} \sum_{l=1}^{L} (A^*_t)_i
\label{eq:4}
\end{equation}
The OAL ensures that the generated objects are precisely positioned according to the input segmentation maps. The loss is computed as a sum of binary cross-entropy losses (\(\mathcal{L}_{\text{BCE}}\)) for each segment, comparing the averaged attention maps to the input segmentation mask:
\begin{equation}
\mathcal{L}_{\text{OAL}} = \sum_{}^{} \left( \mathcal{L}_{\text{BCE}}(\hat{S}_i, S_i) + \mathcal{L}_{\text{BCE}}\left(\frac{\hat{S}_i}{\|\hat{S}_i\|_\infty}, S_i\right) \right).
\label{eq:5}
\end{equation}
\textbf{Discussion 1: }OAL addresses two critical challenges in object replacement:
(a) Attention Dilution: By averaging attention maps across layers/heads, weak localization signals from individual attention heads are aggregated into coherent object masks. This combats the diffusion model's tendency to spread attention broadly.
(b) Scale Invariance: The dual BCE terms balance absolute alignment and scale-normalized alignment. For example, in Fig. \ref{fig:3} lower, this ensures the “dog” token's attention precisely matches the cat's mask without being skewed by residual “red hat” attention.

\subsubsection{Color Consistency Loss (CCL)}
The Color Consistency Loss (CCL) ensures that the colors of edited objects not only align with the target text prompt but also accurately correspond to specified regions in the target image. The CCL maximizes the cross-attention value of the edit token within the masked region, ensuring that the color change is localized and consistent with the desired prompt. The loss is defined as:
\begin{equation}
\mathcal{L}_{\text{CCL}} = \left(1 - S \odot \sum  \frac{(A_t^*)_i}{(A_t^*)_i + \sum_{j=1}^{J}(A_t)_j}\right)^2,
\label{eq:6}
\end{equation}
where \((A_t^*)_i\) is the cross-attention map of the new editing token (\textit{e.g.}, “red” in Fig. \ref{fig:2}), and \((A_t)_j\) represents the cross-attention maps of common tokens from the reconstruction branch.

\noindent \textbf{Discussion 2: }CCL operates through three key mechanisms:
(a) Selective Attention Boosting: By dividing edit token attention by the sum of all token attention within mask \( S \), the loss forces the model to prioritize new attributes over existing content.
(b) Gradient Amplification: The squared term penalizes deviations quadratically, making the model highly sensitive to color misalignments.
(c) Spatial Focus: The mask \( S \) restricts the loss to editing regions, ensuring color changes do not bleed into adjacent objects.

\subsection{Latent Variable Optimization} \label{sec:4.3}
In the Masked Dual-Edit framework, latent variable optimization aims to control both object structure and attribute while preserving the overall image integrity. The total loss is composed of the Object Alignment Loss (OAL) and the Color Consistency Loss (CCL):
\begin{equation}
\mathcal{L}_{\text{total}} = \lambda_1\mathcal{L}_{\text{OAL}} + \lambda_2 \mathcal{L}_{\text{CCL}}.
\label{eq:7}
\end{equation}
where \(\lambda_1\) and \(\lambda_2\) are hyperparameters balancing the structural alignment and appearance consistency objectives, respectively.
The gradient of this total loss is used to update the latent variables \( \mathbf{z} \) through gradient guidance, focusing on the editing regions.
To ensure edits are confined to the specified regions, the latent variables for both editing and non-editing areas are merged using the following equation:
\begin{equation}
\mathbf{z}_t^* = M \odot (\mathbf{z}_t^* - \delta \nabla_{\mathbf{z}_t^*} \mathcal{L}) + (1 - M) \odot \mathbf{z}_t^*,
\label{eq:8}
\end{equation}
where
\( M \) is the mask for all editing regions,
\( \mathbf{z}_t^* \) represents the updated latent variables for the editing regions,
\( \delta \) is the step size for the gradient update.
This process ensures that only the target regions are modified, while the non-editing areas are preserved, enabling localized and precise multi-object editing.

\section{Experiments}

\begin{figure*}
    \centering
    \includegraphics[width=0.9\linewidth]{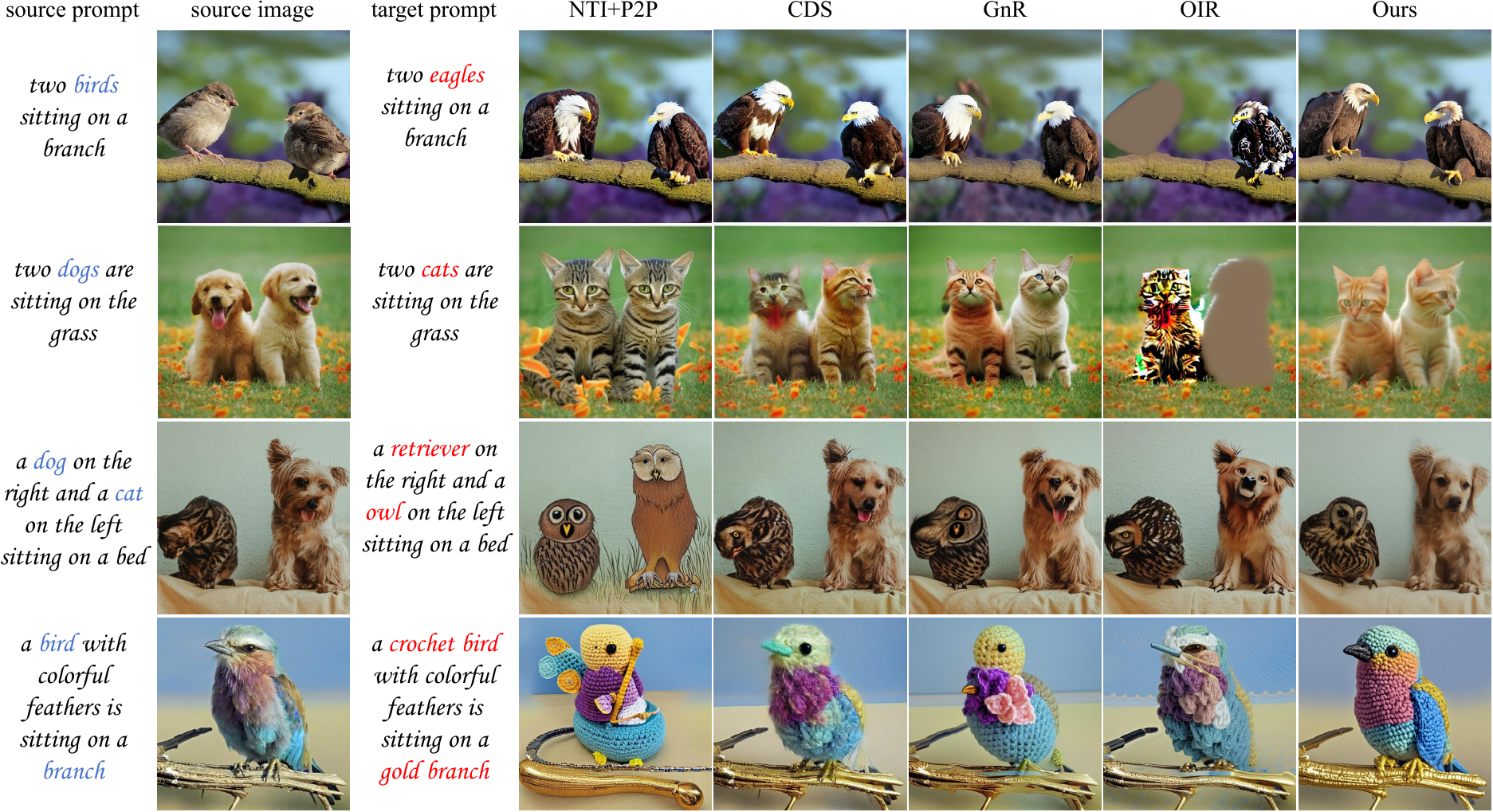}
    \caption{Qualitative comparison with the state-of-the-arts in simple multi-object scenes without overlap. While other approaches struggle with incomplete edits, unintended modifications, or deviations from the prompt, MDE-Edit reliably edits the image as specified.}
    \label{fig:4}
\end{figure*}

\subsection{Experimental Details}
We experimentally evaluate MDE-Edit over two typical datasets, including OIR-bench\cite{yang2023object} and LoMOE-bench\cite{chakrabarty2024lomoe}. The two datasets contain complex images with multiple objects, enabling a more comprehensive evaluation of our method in real-world editing scenarios. Additionally, we select images from MS-COCO\cite{lin2014microsoft} that include multiple objects, ensuring a diverse range of indoor and outdoor scenes. The corresponding edit masks are generated using the SAM (Segment Anything Model)\cite{kirillov2023segment}, which allows precise segmentation for targeted edits. These datasets are used to evaluate localized edits for color, texture, and object replacement tasks. We perform multi-object image editing by optimizing the overall objective in Eq. \eqref{eq:8} during the initial 20 timesteps to achieve precise text-to-image alignment for multiple targets.
Through empirical tuning across multiple trials, we set \(\lambda_1\) and \(\lambda_2\) to 1 and 1.25, respectively, to strike a careful balance between preserving structural integrity and enabling nuanced appearance modifications in multi-object edits. We use Stable Diffusion v1.4 as the backbone model; the denoising process consists of 50 timesteps in total, and all experiments are conducted on a single NVIDIA A6000 GPU. Our code can be accessed in the supplementary material.

\vspace{-4pt} 
\subsection{Comparison with State-of-the-arts}
To validate the superiority of MDE-Edit, we conduct comprehensive comparisons with state-of-the-art image editing methods including OIR\cite{yang2023object}, which performs multi-object editing through a disassembly-reassembly strategy, CDS\cite{nam2024contrastive} that employs contrastive denoising scores with self-attention features to maintain structural consistency, Guide-and-Rescale (GnR)\cite{titov2024guide} which utilizes self-guidance and noise rescaling for tuning-free global editing, and Null-text Inversion+P2P (NTI+P2P)\cite{mokady2023null, hertz2022prompt} combining null-text inversion with cross-attention control for prompt-aligned edits.
We evaluate our method using three key metrics: CLIP Score\cite{hessel2021clipscore}, which measures the alignment between the generated image and the target text prompt; BG-LPIPS\cite{zhang2018unreasonable}, which assesses the perceptual similarity in the background regions between the original and edited images; and BG-SSIM\cite{wang2004image}, which quantifies the structural similarity of the background areas. These metrics together help ensure that the edits are semantically accurate while preserving the image's background and structure.

\begin{figure*}
    \centering
    \includegraphics[width=0.9\linewidth]{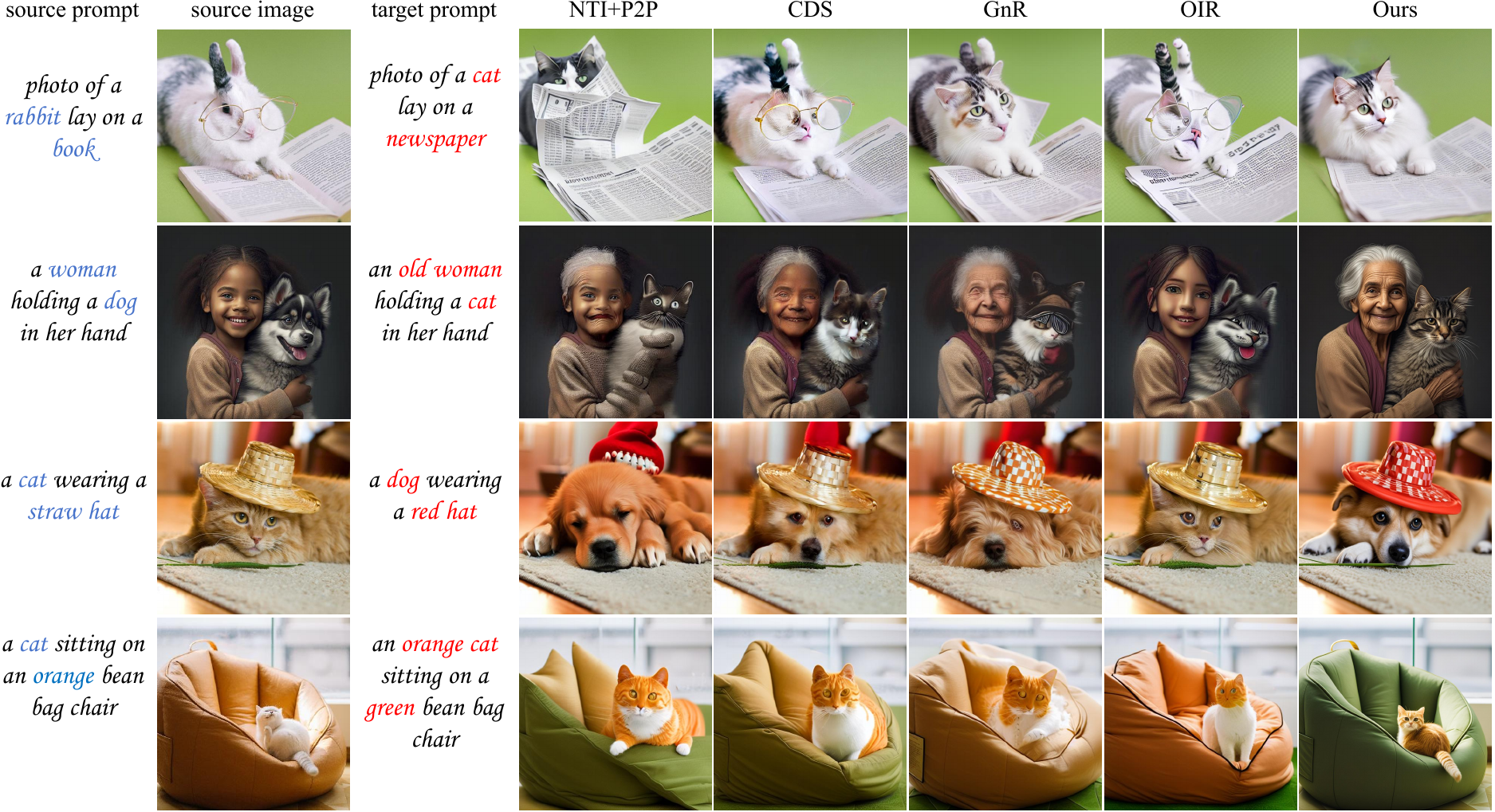}
    \caption{Qualitative comparison with state-of-the-art methods in multi-object overlapping scenarios. In complex scenarios involving multi-object overlap, existing methods often produce suboptimal results—either failing to modify the intended regions, distorting overlapping elements, or introducing inconsistencies with the target description. MDE-Edit, however, demonstrates exceptional precision in such challenging conditions, seamlessly applying edits only to specified objects while preserving overlapping and adjacent areas without interference.}
    \label{fig:5}
\end{figure*}

\begin{table*}[htbp]
  \centering
  \caption{Comparison of quantitative results with SOTA for Multi-Object Edits. $\uparrow$: Higher is better; $\downarrow$: Lower is better. The best and second-best results are reported with \textbf{boldface} and \underline{underline}, respectively.}
  \label{tab:example}
  \begin{tabular}{c c ccc ccc}
    \toprule
    \multirow{2}{*}{Method} & \multirow{2}{*}{Pub'year} & \multicolumn{3}{c}{Multi-object Non-Overlap} & \multicolumn{3}{c}{Multi-object Overlap} \\
    \cmidrule(lr){3-5} \cmidrule(lr){6-8}
     & & CLIP SCORE $\uparrow$ & BG-LPIPS $\downarrow$ & BG-SSIM $\uparrow$ & CLIP SCORE $\uparrow$ & BG-LPIPS $\downarrow$ & BG-SSIM $\uparrow$ \\
    \midrule
    NTI+P2P & CVPR'23 & 0.258 & 0.217 & 0.820 & \underline{0.267} & 0.149 & 0.842  \\
    CDS & CVPR'24 & 0.262 & 0.201 & 0.778 & 0.223 & 0.261 & 0.765  \\
    GnR & ECCV'24 & \underline{0.278} & 0.229 & 0.773 & 0.255 & 0.185 & 0.822  \\
    OIR & ICLR'24 & 0.255 & \underline{0.111} & \underline{0.883} & 0.248 & \underline{0.147} & \underline{0.847}  \\
    \midrule
    Ours & - & \textbf{0.282} & \textbf{0.106} & \textbf{0.925} & \textbf{0.290} & \textbf{0.086} & \textbf{0.936} \\
    \bottomrule
  \end{tabular}
  \label{tab:1}
\end{table*}

\noindent \textbf{Qualitative and Quantitative Results.}
The experimental results, supported by comprehensive qualitative and quantitative analyses, demonstrate the superior performance of MDE-Edit in multi-object image editing across diverse scenarios, ranging from non-overlapping simple cases to overlapping complex instances.

\begin{figure}
    \centering
    \includegraphics[width=1\linewidth]{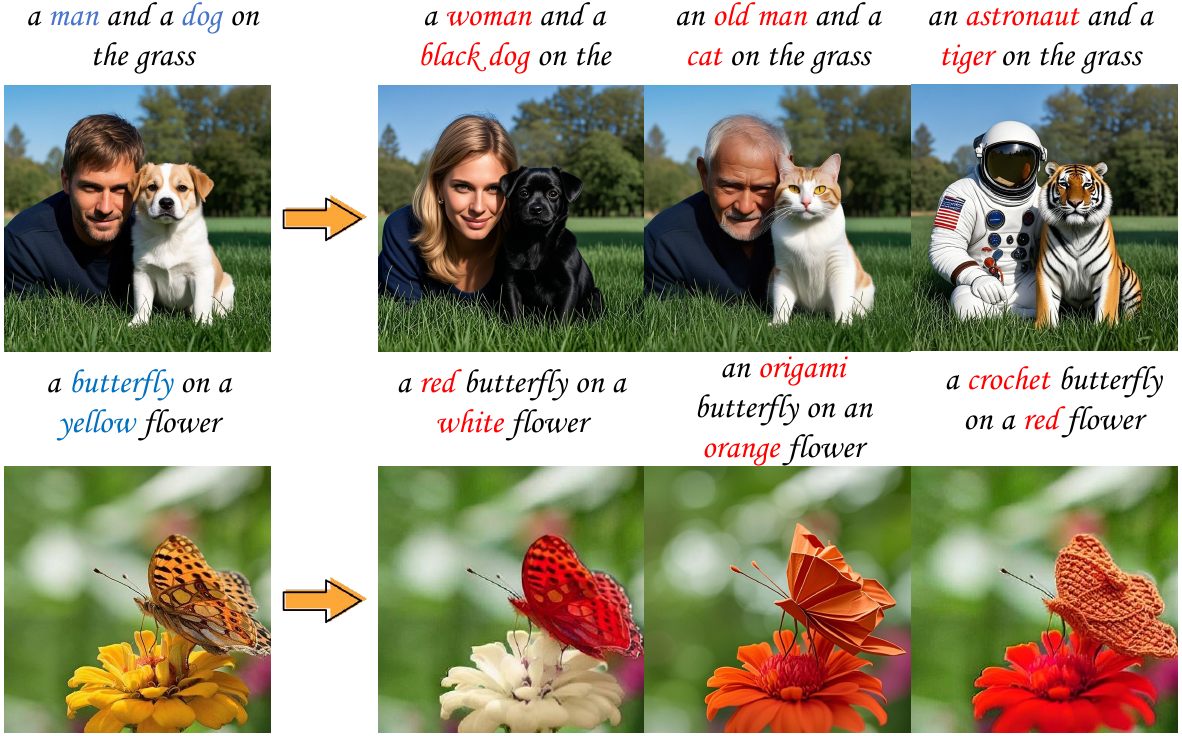}
    \caption{Multi-object editing results of MDE-Edit. Showcasing the diverse multi-object editing capabilities of MDE-Edit, featuring varied subjects (humans, animals, crafted objects) and contexts (natural scenes, imaginative scenarios), highlighting the method's adaptability across styles and compositions.}
    \label{fig:6}
\end{figure}

Qualitatively, MDE-Edit addresses key limitations of existing methods across multi-object scenarios, showcasing significant advantages in both simple non-overlapping and complex overlapping settings. In non-overlapping scenes, such as those depicted in Fig. \ref{fig:4}, it precisely modifies target objects while preserving the surrounding context with remarkable accuracy, achieving seamless texture and color edits devoid of artifacts. Baseline methods struggle to match this level of precision: OIR often results in blurred details due to its reliance on segmentation masks; CDS tends to affect neighboring regions inadvertently, leading to unintended modifications; GnR lacks the spatial accuracy necessary for fine-grained control; and NTI+P2P shows inconsistency in attribute changes. By contrast, MDE-Edit’s mask-guided strategy ensures localized edits are performed without compromising the structural integrity of the image.
In complex overlapping scenarios illustrated in Fig. \ref{fig:5}, MDE-Edit excels by effectively resolving feature entanglement, a major hurdle for competing approaches. Unlike OIR and CDS, which fail to disentangle overlapping features accurately, resulting in ghosting effects or unwanted color bleeding, MDE-Edit maintains clarity and distinction between overlapping entities. GnR suffers from structural distortions in areas of heavy overlap, whereas NTI+P2P exhibits unstable performance under similar conditions. Leveraging Object Alignment Loss (OAL) and Color Consistency Loss (CCL), MDE-Edit aligns attention maps with segmentation masks to enforce strict spatial confinement, while simultaneously suppressing interference from shared tokens through targeted enhancement of edit-specific attributes within designated regions. As demonstrated in Fig. \ref{fig:6}, this dual approach adeptly manages a wide range of overlapping patterns—from partial to highly intricate—preserving not only the finest details but also ensuring semantic consistency and visual harmony throughout the edited images. Collectively, these enhancements enable MDE-Edit to outperform baseline methods significantly in terms of localization and coherence, establishing itself as a robust solution for diverse multi-object editing tasks.

The quantitative evaluation in Table \ref{tab:1} conclusively demonstrates the superiority of MDE-Edit in both simple multi-object scenes without overlap and complex multi-object scenes with overlap. Across all key metrics, MDE-Edit outperforms existing methods by effectively addressing their core limitations. In non - overlapped tasks, it achieves superior text - image alignment and background fidelity, overcoming issues such as incomplete edits (\textit{e.g.}, CDS). For overlapped scenarios, MDE - Edit maintains exceptional precision even in complex overlapping regions, where baseline methods like GnR and NTI+P2P suffer from attention misalignment and attribute conflicts (\textit{e.g.}, hybrid artifacts or color leakage). While OIR demonstrates competitive performance on BG-LPIPS and BG-SSIM metrics, it fails to reliably recognize multiple instances of the same object in non-overlapped multi-object scenarios. Moreover, in overlapped multi-object editing tasks, OIR struggles to maintain robust performance.

Collectively, the analyses demonstrate the critical and complementary roles of OAL and CCL: OAL achieves precise object localization through its mask-aligned attention mechanism, whereas CCL ensures consistent attribute binding within designated regions.
This dual strategy decouples structural and appearance editing, addressing core multi-object challenges to enable superior semantic and visual coherence in complex scenes.

\begin{table}[h]
\centering
\caption{Ablation experiments in multi-object editing. $\uparrow$: Higher is better; $\downarrow$: Lower is better. The best results are reported with \textbf{boldface}.}
\label{tab:performance}
\begin{tabular}{lccccc}
\toprule
\multirow{3}{*}{Setting} & \multirow{3}{*}{$\mathcal{L}_{\text{OAL}}$} & \multirow{3}{*}{$\mathcal{L}_{\text{CCL}}$} &
\multicolumn{3}{c}{Evaluation Metrics} \\
\cmidrule(lr){4-6}
 & & & CLIP & BG-LPIPS & BG-SSIM \\
 & & & SCORE $\uparrow$ & $\downarrow$ & $\uparrow$ \\
\midrule
Setting 1 & $\times$ & $\times$ & 0.239 & 0.219 & 0.796 \\
Setting 2 & $\checkmark$ & $\times$ & 0.265 & 0.153 & 0.850 \\
Setting 3 & $\times$ & $\checkmark$ & 0.253 & 0.188 & 0.818 \\
Setting 4 & $\checkmark$ & $\checkmark$ & \textbf{0.284} & \textbf{0.095} & \textbf{0.918} \\
\bottomrule
\end{tabular}
\label{tab:2}
\end{table}

\begin{figure}
    \centering
    \includegraphics[width=1\linewidth]{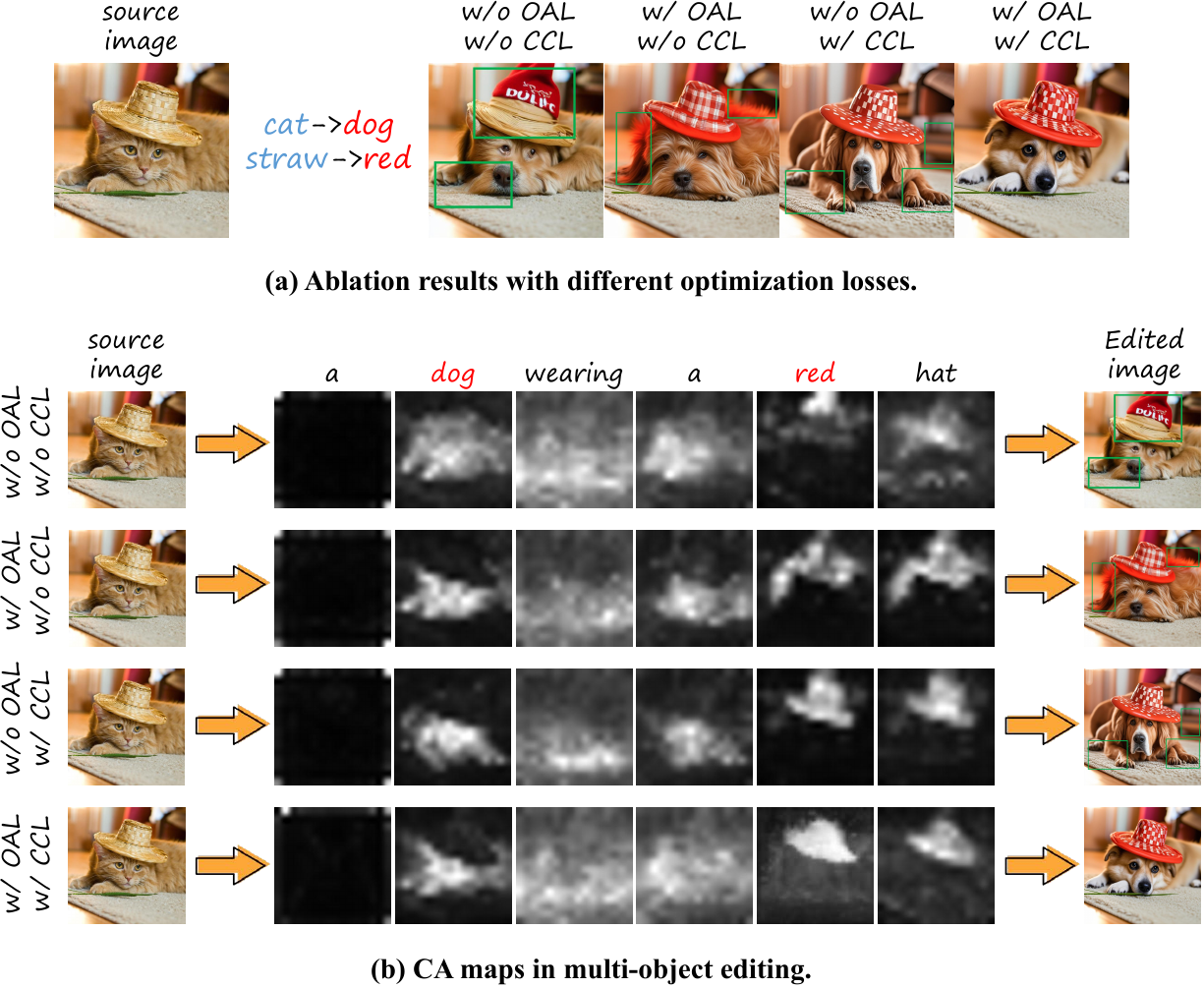}
    \caption{Ablations for MDE-Edit. (a) Ablation results in multi-object editing with different optimization losses. (b) Examples of CA maps in multi-object editing.}
    \label{fig:7}
\end{figure}

\subsection{Ablation Studies}
The ablation studies, as demonstrated in Table \ref{tab:2} and Fig. \ref{fig:7}, rigorously dissect the distinct contributions of the Object Alignment Loss (OAL) and Color Consistency Loss (CCL) across experimental configurations. In Setting 1 (no losses applied), the absence of both OAL and CCL results in pronounced attention misalignment and semantic incoherence, exemplified in Fig. \ref{fig:7}(a) column 1 by hybrid artifacts (\textit{e.g.}, partial object replacements) and unconstrained attribute diffusion (\textit{e.g.}, unintended color bleeding into non-target regions). Setting 2 (OAL-only) addresses structural challenges by aligning cross-attention maps with segmentation masks, achieving precise geometric localization (\textit{e.g.}, correct spatial placement of a "cat" replacement in Fig. \ref{fig:7}(a) column 2). However, without CCL, attribute leakage persists, as color modifications fail to adhere strictly to masked regions. Conversely, Setting 3 (CCL-only) enforces localized attribute consistency through selective attention amplification (\textit{e.g.}, confining "red" to the hat region in Fig. \ref{fig:7}(a) column 3), yet lacks structural constraints, leading to boundary ambiguities (\textit{e.g.}, distorted object contours) and positional mismatches.
The synergistic integration of OAL and CCL in Setting 4 resolves these limitations: OAL anchors edits to geometrically accurate regions via mask-aligned attention (\textit{e.g.}, preserving the original scale and position of replaced objects in Fig. \ref{fig:7}(a) column 4), while CCL suppresses cross-attention leakage by amplifying target attributes within masks (\textit{e.g.}, isolating "red" to the hat without affecting adjacent regions). This dual mechanism disentangles structural preservation and attribute binding, effectively mitigating attention interference between multiple objects (\textit{e.g.}, disentangling "cat" and "red hat" in complex scenes) and latent feature conflicts (\textit{e.g.}, color/texture entanglement). Fig. \ref{fig:7}(b) further illustrates the individual impacts of the two losses by visualizing cross-attention maps during editing. It shows how OAL sharpens attention localization to match object masks (enhancing structural precision) and how CCL boosts target attribute attention within masked regions while suppressing irrelevant token interactions (improving attribute consistency). These visualizations corroborate the quantitative results, highlighting the complementary roles of OAL and CCL in achieving coherent multi-object edits, validating their joint optimization is essential for addressing localization and attribute mismatch challenges.

\section{Conclusion}
In this work, we propose MDE-Edit, an advanced framework for multi-object image editing that integrates Cross-Attention Control and a Masked Dual-Edit strategy. By combining Object Alignment Loss (OAL) and Color Consistency Loss (CCL), our method achieves precise control over object structure and appearance, effectively addressing challenges of inaccurate target object localization and attribute-object mismatch in complex scenes. Experimental results validate the superiority of MDE-Edit over existing methods, showcasing its ability to deliver high-fidelity edits in diverse multi-object scenarios. Our framework advances the application of diffusion models in image editing, providing a scalable and efficient solution for real-world tasks.


\bibliographystyle{ACM-Reference-Format}
\bibliography{sample-base}










\end{document}